\documentclass[authoryear]{elsarticleGeneric}

\usepackage{amssymb}
 \usepackage{amsthm}

 \usepackage{amsmath}
 \usepackage{hyperref}
 \usepackage{booktabs}
\usepackage{lineno}
\journal{Neural Networks}

\begin{document}

\begin{frontmatter}



\title{Putting a bug in ML: The moth olfactory network learns to read MNIST}


\author[eeuw]{Charles B. Delahunt\corref{cor1}} % \fnref{fn1} }
\ead{delahunt@uw.edu}
\author [amuw]{J. Nathan Kutz\corref{cor2}} % \fnref{fn2} }
\ead{kutz@uw.edu}
\cortext[cor1]{Principal corresponding author  $~~~~~~~~~~~~~~~~~~~~~~~~~~~~~~~~~~~~~~~~~~~~~~~$v3,  25 January 2019}
 \cortext[cor2]{Corresponding author}
\address[eeuw]{Department of Electrical Engineering, University of Washington, Seattle\\
Computational Neuroscience Center, University of Washington, Seattle}
\address[amuw]{Department of Applied Mathematics, University of Washington, Seattle}

\address{}

\begin{abstract}

We seek to (i) characterize the learning architectures exploited in biological neural networks for training on very  few samples, and (ii) port these algorithmic structures to a machine learning context. 
The Moth Olfactory Network  is among the simplest biological neural systems that can learn, and its architecture includes key structural elements and mechanisms widespread in biological neural nets, such as cascaded networks, competitive inhibition, high intrinsic noise, sparsity, reward mechanisms, and Hebbian plasticity. 
These structural biological elements, in combination, enable rapid learning.

MothNet is a computational model of the Moth Olfactory Network, closely aligned with the moth's known biophysics and with \textit{in vivo} electrode data  collected from moths learning new odors.
We assign this model the task of learning to read the MNIST digits. 
We show that MothNet successfully learns to read given very few training samples (1 to 10 samples per class). 
In this few-samples regime, it  outperforms standard machine learning methods such as nearest-neighbors, support-vector machines, and neural networks (NNs), and matches specialized one-shot transfer-learning methods but without the need for pre-training.

The MothNet architecture  illustrates how algorithmic structures derived from biological brains can be used to build alternative NNs that may avoid some of the learning rate limitations of current engineered NNs.
%
% This novel, bio-inspired architecture points to a potentially valuable, complementary approach to current NN design.
%	
\end{abstract}

\begin{keyword}
machine learning \sep Hebbian \sep sparsity  \sep olfactory network  \sep bio-mimesis  \sep neural networks  \sep one-shot learning
\end{keyword}

\end{frontmatter}

%\clearpage

%--------------------------------------------------------------------------------------------------------------------------------------------------------------
%---------------------------------------------------------                   INTRODUCTION                         -------------------------------------------
%--------------------------------------------------------------------------------------------------------------------------------------------------------------

% \linenumbers

\section{Introduction}
\label{introduction}

Originally inspired by the biological structure of networks of interacting neurons [\cite{mcCulloch1943, hubel, rosenblatt1958, fukushima}], neural networks (NNs) have since developed a suite of algorithmic tools (such as backprop, convolutional kernels, etc) which, combined into complex and deep NN architectures, have achieved unprecedented success in a wide array of machine learning (ML) tasks [\cite{schmidhuber2014, goodfellow2016deep, lecun2015deep}].  
However, they have trouble learning from few samples.  
We seek to improve NN performance on such tasks by revisiting the well of biological example to characterize key biological structures involved in learning, for transfer to the deep NN context.  

In this work, we apply MothNet, a NN architecture closely based on the \textit{Moth Olfactory Network}, to the task of learning vectorized MNIST digits.
This moth-based architecture successfully learns to read given very few training samples (1 to 10 samples per class) and in fact outperforms standard machine learning methods such as Nearest-Neighbors, support-vector machines (SVMs), Neural Nets (i.e. standard fully-connected, feed-forward, non-convolutional nets with softmax loss functions), and convolutional neural networks (CNNs), in the few-sample regime (Fig \ref{accuracyVariousMethods}). 
In addition, it performs comparably to or better than specialized one-shot learning NNs, but without their need for pre-training to initialize NN weights (Table \ref{table1}).

The Moth Olfactory Network is among the simplest biological neural networks (BNNs) that can learn [\cite{riffell2013}] \footnote{References  include studies of various insect species, which we use subject to the constraint that inter-species tranferability varies (the fly and moth are similar;  while locust and honeybee results are sometimes general enough to transfer, sometimes not). MothNet is based on the moth olfactory network in particular.}.  It is well-characterized, and it contains key features widespread in BNNs, including  high noise [\cite{galizia2014}], random connections [\cite{caron2013}], competitive inhibition [\cite{hong2015}], Hebbian synaptic growth [\cite{cassenaer, masse2009}], high-dimensional sparse layers [\cite{campbellMushroomBody, honeggerTurner2011}], large dimension shifts between layers [\cite{babadi}], and generalized stimulation of neurons during learning  [\cite{hammerMenzel1998}].  
It thus offers an ideal avenue to investigate biological learning and the interacting components that make learning possible. 
 
The olfactory processing unit centers on two interacting networks, the noisy antennal lobe (AL) and the sparse mushroom body (MB), known as the AL-MB  [\cite{wilson2008, campbellMushroomBody}].  
The AL pre-processes odor inputs, then projects the odor signal forward to the MB, which in turn feeds forward to readout neurons.
Learning occurs when the neuromodulatory chemical octopamine (triggered by sucrose reward) induces an overall increase in excitation in the AL [\cite{dacksRiffell2009}], causing the plastic MB synapses to update via Hebbian (``fire together, wire together'') growth. 

A computational model of a stand-alone honeybee MB (i.e. without the AL or octopamine dynamics) was trained by [\cite{huertaNowotny2009}] on the handwritten vectorized MNIST digits dataset [\cite{leCunMnist}].
The trained honeybee MB attained over 80\% accuracy given sufficient training samples (200 to 2000 samples per class), indicating that the MB is to some degree task-agnostic, i.e. not limited solely to odor processing.

MothNet is an end-to-end computational model of the {\em Manduca sexta} (Hawk moth) olfactory network developed by [\cite{delahuntMoth1}] to study the moth's olfactory learning mechanisms.
MothNet's architecture is closely based on the moth's known biophysical structure. 
It includes the AL, the MB, and readout neurons, as well as Hebbian plasticity and neuromodulatory stimulation by octopamine.
Its behavior is calibrated to, and consistent with,  \textit{in vivo} firing rate data recorded in moths learning new odors. 
For full details see [\cite{delahuntMoth1}]. 

To test the generalizability of this learning architecture, we assigned MothNet  the ML task of identifying downsampled, vectorized MNIST digits (hereafter ``$v$MNIST'' to emphasize that the inputs lack spatial structure).
We made minimal modifications to the MothNet architecture: We simply replaced odor inputs to the AL neurons with pixel values, increased the number of AL units from 60 to 85, and attached 10 readout neurons, one for each $v$MNIST digit.
MothNet routinely achieved 70\% to 80\% accuracy classifying test digits after training on 1 to 10 samples per class, 
out-performing standard ML methods.
These results demonstrate that even a very simple biological architecture contains novel and effective tools that are useful for ML tasks, in particular tasks constrained by limited training data or the need to add and train new classes without retraining the full NN.

%--------------------------------------------------------------------------------------------------------------------------------------------------------------
%------------------------------------------------         MOTH OLFACTORY NETWORK                                    -----------------------------
%--------------------------------------------------------------------------------------------------------------------------------------------------------------

\section{The moth olfactory network and MothNet model}
\label{mon}

\subsection{Moth olfactory network outline}  

%This stimulated AL activity induces Hebbian updates to plastic synaptic weights in the plastic MB, i.e. modulating one network trains another.  
%Sparsity in the MB controls Hebbian plasticity by filtering noise and by focusing training-induced weight updates on relevant signals. 
%In combination, these features enable rapid and effective learning, expressed as permanent modulation of readout neuron responses to trained odors. \\\\
 
\begin{figure}[t]
\centering
\fbox {
\includegraphics  [width=0.96\linewidth]{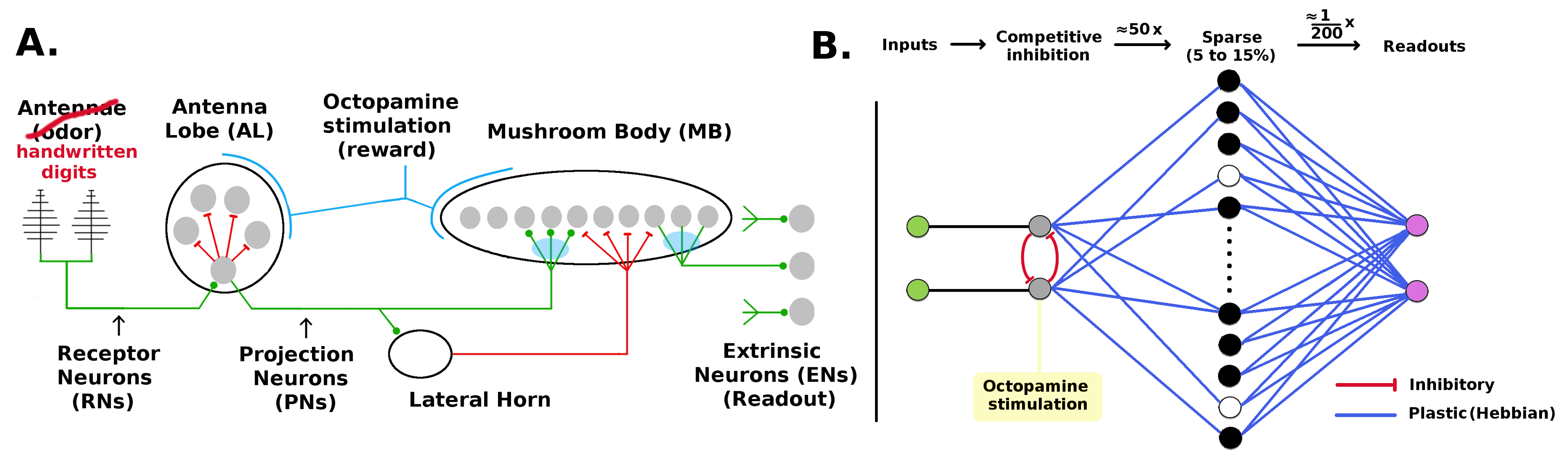} 
}
\caption{\small { \textbf{Network schematics.} 
{\bf{A:}}  Biological schematic. Green lines show excitatory connections, red lines show inhibitory connections.
Light blue ovals show plastic connections into and out of the MB. 
The units in the AL competitively inhibit each other. 
Global inhibition from the lateral horn induces sparsity on MB responses. 
The ENs give the final, actionable readouts of the system's response to a stimulus.
{\bf{B:}}  Alternative schematic of the same network.
} }
\label{monSchematic}
\end{figure}

The network   is organized as a feed-forward cascade of five distinct networks, as well as a reward mechanism  [\cite{martin2011, kvello2009}]. 
Fig~\ref{monSchematic} gives a system schematic.

Starting at the Antennae, several thousand noisy chemical Receptor Neurons (RNs) detect odor and send signals to the Antennal Lobe (AL) [\cite{wilson2008, masse2009}].
The AL acts as a pre-amp, providing gain control and sharpening odor codes through lateral inhibition [\cite{bhandawat2007}].
The AL contains roughly 60 isolated units (glomeruli), each focused on a single feature; 
that is, all antennae receptors for a particular odor volatile converge onto a single dedicated AL unit [\cite{martin2011}].
These units contain various types of neurons.
The AL projects odor codes forward to the Mushroom Body (MB) [\cite{campbellMushroomBody}],  in excitatory Projection Neurons (PNs) that randomly connect (non-densely) to MB neurons [\cite{caron2013}].

The MB contains about 4000 Kenyon Cells, which fire sparsely and  encode odor signatures as memories [\cite{perisse2013, honeggerTurner2011}] (we use ``MB'' as a synonym for these Kenyon Cells).
Sparsity in the MB is enforced by global inhibition from the Lateral Horn [\cite{bazhenovStopfer2010}].
The MB  feeds forward to Extrinsic Neurons (ENs), numbering $\sim$10's, which are ``readout neurons" that interpret the MB codes  and deliver actionable output to the rest of the moth [\cite{campbell2013, hige2015}].

The network can learn:
In response to reward (sugar at the proboscis), a large neuron releases the neuromodulator octopamine globally over the AL and MB.
Octopamine stimulates AL neurons, inciting higher firing rate responses to the reinforced stimuli.
This in turn induces growth in the plastic synaptic connections into the MB (AL$\rightarrow$MB) and out of the MB (MB$\rightarrow$ENs) via Hebbian updates  [\cite{cassenaer, masse2009}]. 
Learning does not occur without this octopamine input  [\cite{hammer1995, hammerMenzel1998}]. 

% Firing rate (FR) timecourses for neurons in each network, from a typical MothNet training simulation, are shown in Fig~\nameref{schematicPlusTimecourses} .

\subsection{MothNet model}
The MothNet computational model closely follows this biological architecture in terms of connections, numbers of neurons in each layer, etc [\cite{delahuntMoth1}]. 
Neural firing rates are modeled with firing-rate dynamics [\cite{dayan2001}] evolved as stochastic differential equations  (SDEs) [\cite{higham2001}]:
\begin{equation}
\tau \dfrac{dx}{dt} =  -x + s (\Sigma {\bf{w}}_{i} {\bf{u}}_{i} ) = -x + S({\bf{w}\cdot\bf{u}}) +  dW , % \text{ where}
\end{equation}
where $x(t)$ = firing rate (FR) for a neuron; 
{\bf{w}} = connection weights;
{\bf{u}} = upstream neuron FRs;
$S()$ is a sigmoid function or similar; and
$W(t)$ = a brownian motion process.

In MothNet this equation is modified by an inserted term to model the stimulative effect of octopamine on the AL during learning, since this effect is central to learning in the actual moth [\cite{delahuntMoth1}].
The model uses a simple model of Hebbian plasticity for synaptic weight updates [\cite{hebb, dayan2001,roelfsema2018}]:
\begin{equation} \label{hebbianGrowthEqn}
\Delta w_{ab}(t) = \gamma f_a(t) f_b(t) 
\end{equation}
where $f_a(t), f_b(t)$ are firing rates of neurons $a, b$ at time $t$; $w_{ab}$ is the synaptic weight between them; and $\gamma$ is a growth rate parameter.\\
In addition, inactive  MB$\rightarrow$EN weights are subject to proportional decay: 
\begin{equation} \label{decayEqn}
\Delta w_{ab}(t) = -\delta w_{ab}(t) \text{,  if }  f_a(t) f_b(t) = 0. 
\end{equation}
where $\delta$ is a decay parameter.
There are two layers of plastic synaptic weights:  AL$\rightarrow$MB, and MB$\rightarrow$ENs (i.e. pre- and post-MB).
Hebbian plasticity is assumed to be ``switched on'' by reward, so it is cosynchronous with octopamine.
Thus, plasticity only occurs during training sessions.
There is no built-in stabilization of weight updates, except the upper and lower (i.e. 0) rails of allowable weights.
As seen in section \ref{sparsity}, sparsity in the MB is crucial to prevent Hebbian weight updates from running out of control and amplifying noise channels.

For the $v$MNIST task, MothNet has 10 extrinsic neurons (ENs), that start with identical MB$\rightarrow$EN connections.
When training begins, these 10 ENs are randomly assigned to target particular digits 0 to 9.
Once ENs are assigned, training is supervised: When a digit of class \textit{j} is presented, only MB$\rightarrow$EN$_j$ connections are updated, where EN$_j$ is the EN assigned to class \textit{j}.
That is, the system knows the class of the training sample, for the purposes of post-MB updates.
Training rapidly individualizes these weights according to their target digits.
In contrast, all pre-MB connections are updated in every case, since these connections are common to all inputs (in these experiments, pre-MB connections were essentially fixed due to slow learning rates).

We note that in MothNet, updates do not follow the same logic as in typical ML algorithms: 
There is no objective function to be minimized, and no output-based loss that is pushed back through the network  as with backprop or agent-based reinforcement learning (there is no ``agent'' in the MothNet system).
The Hebbian weight updates, either growth or decay, occur on a purely local ``use it or lose it'' basis.

% A rough balance between growth and decay rates is important for optimal learning.
% Full governing equations are given in \ref{fullDynamics}.

%----------------------------------------------------------------------------------------------------------------------

%--------------------------------------------------------------------------------------------------------------------------------------------------------------
%------------------------------------------------         METHODS                   -------------------------------------------------------------------------
%--------------------------------------------------------------------------------------------------------------------------------------------------------------

\section{Methods}
\subsection{Training data}
We did not use the MNIST dataset in its usual format of images with spatial structure.
Rather, we derived an 85-feature, 10-class dataset, with no spatial information,  from the MNIST dataset.
MNIST images (from PMTK3 [\cite{murphyML}]) were downsampled and vectorized, and pixel values then served as input features.
Of course both biological and engineered systems routinely choose better feature sets.
%  (for example, convolutional kernels \cite{leCun1989}.  
However, pixels-as-features provided a good test of whether MothNet can effectively learn to discriminate classes given inputs with inter-class correlations.

\begin{figure}[t]
\centering
\fbox {% {\rule[0.5cm]{0cm}{0cm}
\includegraphics  [width=0.96\linewidth]{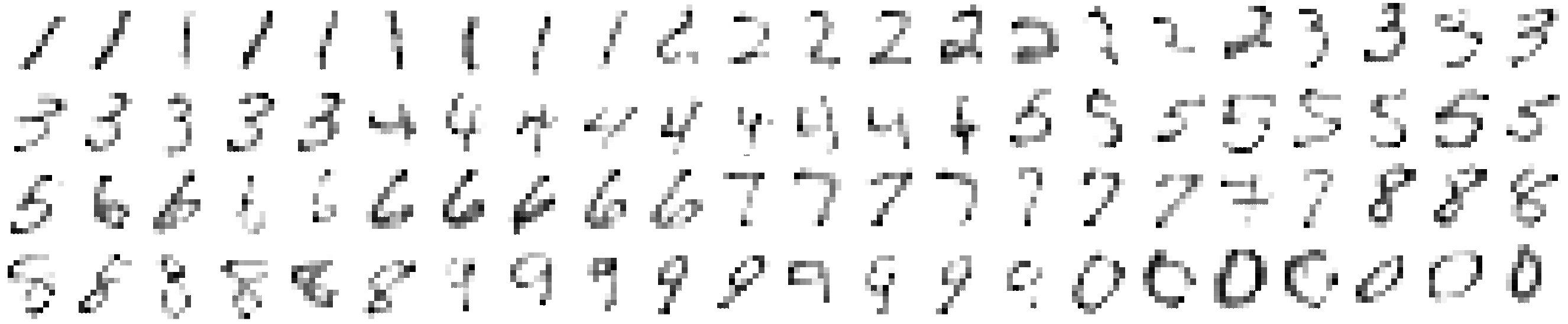} 
%  \rule[0.5cm]{0cm}{0cm}
}
\caption{\small { Cropped, downsampled, mean-subtracted and zero-thresholded  MNIST digit images (pre-processing steps 1 and 2), shown before further sub-selection of globally most-active pixels and vectorization (steps 3 and 4). 
Vectors of these pixels were the feature vectors for the experiments.
} }
\label{typicalMnist}
\end{figure}

MothNet (like the moth) feeds one feature to each unit in the AL. 
Using full MNIST images, vectorized to give pixels-as-features, would imply $28^2 = 784$ units
To keep the scale of MothNet somewhat close to that of the actual moth (60 AL units), we preprocessed the images as follows:\newline
$~~~~~~$1. Crop by 2 and downsample by 2 (linear interpolation), to get 12 x 12 images (shown in Fig \ref{typicalMnist}). \newline
$~~~~~~$2. Mean-subtract using 500 random, set-aside digits (50 from each class) and zero out negative values. \newline
$~~~~~$3. Select the most-active pixels by thresholding the various class averages.
The purpose is to retain the most globally active pixels  while also preserving the most active pixels of each class, and to exclude border pixels which supply little information.
This step  was part of the dataset definition, not part of training, and did not involve the MothNet model.\\
$~~~~~ $4. Vectorize the remaining pixels.

Each retained pixel became a feature that fed into one unit of MothNet's AL.
The experiments described here used 85 pixels (out of $12^2 = 144$ total) to represent the $v$MNIST digits. 
This gave MothNet 85 units in the AL and 2550 neurons in the MB (ie 30 times the number of AL units).
Examples of dowsampled images, before vectorization and restriction to the receptive field pixels, are shown in Fig \ref{typicalMnist}.  \newline

%----------------------------------------------------------------------------------------------------------------------

%----------------------------------------------------------------------------------------------------------------------------
\subsection{Experiment design}
Instances of MothNet are randomly generated from parameter templates.
Particular MothNet behaviors can be modulated by varying one or two template parameters.
Each experiment used a fixed basic template, modified only by varying the parameter(s)-under-test.
Each of these modified templates then randomly generated many moths, typically 13-17 per data point.
The  basic templates varied slightly between experiments, e.g. in some parameter values. 
We found that these slight differences in template had minor effect, and that a wide range of parameters and templates delivered effective learning behavior.\\

%\paragraph*{Experiment sequence of events}
\noindent Each experiment contained three stages:\\  
$~~~$1. Pre-training Baseline (15 digits per class), used to assess naive classifier accuracy; \\ 
$~~~$2. Training ($N$ digits per class, randomly ordered); \\\
$~~~$3. Post-Training Validation (15 digits per class), used to assess post-training classifier accuracy.

Digits for each stage were randomly chosen without replacement from non-intersecting pools. 
Using more than 15 digits in baseline and validation sets did not significantly affect results.

Since the moth olfactory system can learn to recognize a new odor given roughly 8-10 samples [\cite{riffell2013}], we focused on small training sets (1-20 samples/class).
In some experiments, training included multiple ``sniffs'', i.e. repeated presentations of each sample.
The order of training samples did not matter, perhaps because the strongest plasticity was specific to the readout neuron (EN) targeting a given class.

%
%----------------------------------------------------------------------------------------------------------------------------
%
\subsection{Classifiers}  \label{classifier}
System readout units are the Extrinsic Neurons (ENs) downstream from the sparse MB layer and its plastic connections.
These ENs are silent absent any input sample, and they consistently respond, more or less strongly,  to input samples (see Fig~\ref{exampleTimecourse}). 
We use two different classifiers to assess MothNet performance, both dependent on the EN readouts.

\subsubsection{Softmax classifier}
The first method is the basic softmax:
\begin{equation}\label{softmaxEqn}
\hat{s} = \underset{j \in J}{\text{max}} \left\{ \frac{e^{E_j(s)} }{\sum\limits_{i \in J}e^{E_i(s)} } \right\} \text{, where }
\end{equation}
$~~~~~~$$\hat{s}$ = predicted class of sample $s$   \newline
$~~~~~~$$E_i(s)$ = response of the $i$th EN to $s$  \newline
$~~~~~~$$j \in J$ are the classes (0-9).   \newline

This is tantamount to prediction based on the strongest EN response. 
In NNs, the softmax rule is baked into the loss function, so it is a logical classifier by construction.
However,  we see no reason that it would be well-suited to a network that uses Hebbian updates.
In particular, a Hebbian update method does not constrain the scales of the various EN responses  to match each other, as happens in a standard NN with softmax-based loss function.

\subsubsection{Log-likelihood classifier}
We thus use a second classifier as well, which tries to capture the combined statistical content of the various EN readouts without assuming that the EN responses match each other in scale.
This classifier is a summed log-likelihood over the distributions of responses to each digit class in each EN:
\begin{equation} \label{logLikelihoodEqn}
\hat{s} = \underset{j \in J}{\text{min}} \left\{  \sum\limits_{i\in J}\left( \frac{E_i(s) - \mu E_{ij} } { \sigma E_{ij} } \right)^4  \right\} \text{, where }
\end{equation}
$~~~~~~$$\hat{s}$ = predicted class of sample $s$   \newline
$~~~~~~$$E_i(s)$ = response of the $i$th EN to $s$  \newline
$~~~~~~$$\mu E_{ij} = \text{mean} (E_{i}(t) | t \in V, t \in \text{class } j )$   \newline
$~~~~~~$$\sigma E_{ij} = \text{std dev}(E_i(t) | t \in V, t \in \text{class } j )$  \newline
$~~~~~~$$j \in J$ are the classes (0-9) \newline
$~~~~~~$$V$ is a reference set (e.g. a validation set).   \newline  \newline
Roughly, $j$ is a strong candidate for $\hat{s}$ if each EN's response to $s$ is close (in Mahalanobis distance) to that EN's expected response to class $j$.
The use of the $4^{\text{th}}$ power (vs the usual $2^{\text{nd}}$ power) is a sharpener that penalizes outliers.
% If multiple ``sniffs'' were used to test a sample, the EN responses were averaged before calculating $E_i(s), \mu E, \sigma E, \text{and }\hat{s}$. 

This summed log-likelihood is a measure of how well the ENs can separate  the response distributions to various classes, combining information from all ENs and including information about responses to classes not specifically targeted by a particular EN.
The goal of this classifier is to assess how much discriminatory information MothNet was able to extract from the training data. 
We do not wish to imply it is biologically realistic (we don't know).
Accuracy of naive (i.e. untrained) MothNet instances was about 15\% using the log-likelihood classifier, slightly higher than random guessing, perhaps because the digit ``1'' often elicited slightly different naive responses than other digits.

For a given experiment, the post-training classification accuracy was calculated on the validation set, i.e. the same set used to estimate the post-training EN response distribution parameters $\mu E$ and $\sigma E$. 
Similarly, the baseline (pre-training) classification accuracy was calculated on the same baseline set used to estimate naive EN response distribution parameters.
Holdout sets had similar accuracy.

\subsubsection{Pros and cons}
The softmax classifier does not require a validation set, and thus is a more realistic candidate for true few-shot learning, where by definition no validation set is available.
However, it is an arbitrary choice for systems that use a Hebbian learning rule, especially since the Hebbian rule does not ensure a built-in means of scaling EN responses.
Thus softmax may be ill-suited to reflect the full information of the trained network effectively.

The Log-likelihood classifier can arguably claim to better leverage the information contained in a Hebbian-trained network, and it is robust to differences in scale of EN responses.
However, it requires a validation set for its hyperparameters.
In few-shot scenarios, while it may well express the stored learning of the network it is not a practical in-the-field classifier.

We find that in practice, each classifier has strengths. 
In particular, the log-likelihood is stronger given few training samples, while softmax is stronger given many training samples.
We report results for both.

%A more simplistic classifier is simple thresholding: Label the test sample as the class of the most responsive EN$_j$ (by Mahalanobis distance),
%\begin{equation} \label{thresholdEqn}
%\hat{s} = \underset{j \in J}{\text{max}} \left\{ \frac{E_j(s) - \mu E_{jj} } { \sigma E_{jj} }  \right\} \text{, 
%notation as in Eqn.~(\ref{logLikelihoodEqn}). }
%\end{equation}
%This uses no cross-class information ($E_i$ response to $s \in \text{class } i \neq j$).

%Accuracy of classification by simple thresholding was in general much lower than by the log-likelihood classifier, likely because thresholding does not leverage the full information about each EN's responses to every class, but only each EN$_j$'s response to its assigned class $j$. 
%

%--------------------------------------------------------------------------------------------------------------------------------------------------------------
%------------------------------------------------                   RESULTS              ------------------------------------------------------------------------
%--------------------------------------------------------------------------------------------------------------------------------------------------------------

%

\section{Experiments and Results}

This section first presents (i) results of MothNet experiments focused on learning.
It next gives baseline comparisons of MothNet to (ii) standard ML methods, and (iii) specialized ML one-shot methods.
It then presents further results of MothNet experiments focused on (iv) one-shot learning, (v) sniffing, (vi) effects of AL noise, and (vii) MB sparsity.
Computer code for running simulations and experiments in this paper can be found at:\newline 
\url{https://github.com/charlesDelahunt/PuttingABugInML}

\subsection{Learning experiments}
We ran training experiments with various MothNet templates to assess their ability to learn the $v$MNIST digits.
In general, a wide range of templates responded well to training by differentiating their EN responses to different digits (input classes). 
In naive moths, all ENs had the same response profile, and responded similarly to all digit classes, as expected given the symmetry of random connection weights to the various ENs. 
Training caused EN responses to rapidly diverge from baseline and from each other, such that each EN responded most strongly to its assigned digit. 
Common effects of training included: 
\begin{enumerate}
\item Most ENs (e.g. 1, 2 6, 7, 0) tended to amplify the response to their trained digit very well, compared to responses to control digits. 
This gave strong separation and accurate classification.  
\item A few ENs (e.g. 5) sometimes poorly separated their trained digit from control digits. These were the digits most often misclassified during validation.  
\item  Some ENs consistently boosted the responses to certain control digits along with their trained digit (eg, EN$_9$ boosted 4 and 7, EN$_4$ boosted 7 and 9). 
These cases typically reflected visible similarities in the digits, and led to characteristic errors. 
For example, 9s, if misclassified, were often misclassified as 4s.
However, 9's were not misclassified as 7s, because EN$_7$ usually strongly separated 7 from 9. 
That is, outputs of EN$_9$, EN$_7$, and EN$_4$ combined were sufficient to distinguish 9 from 7, but not always 9 from 4.
\end{enumerate}
%----------------------------------------------------------------------------------------------------------------------

\begin{figure}[h!]
\centering
\fbox{ %\rule[.5cm]{0cm}{0cm}  
	\includegraphics [width= 0.96\linewidth] {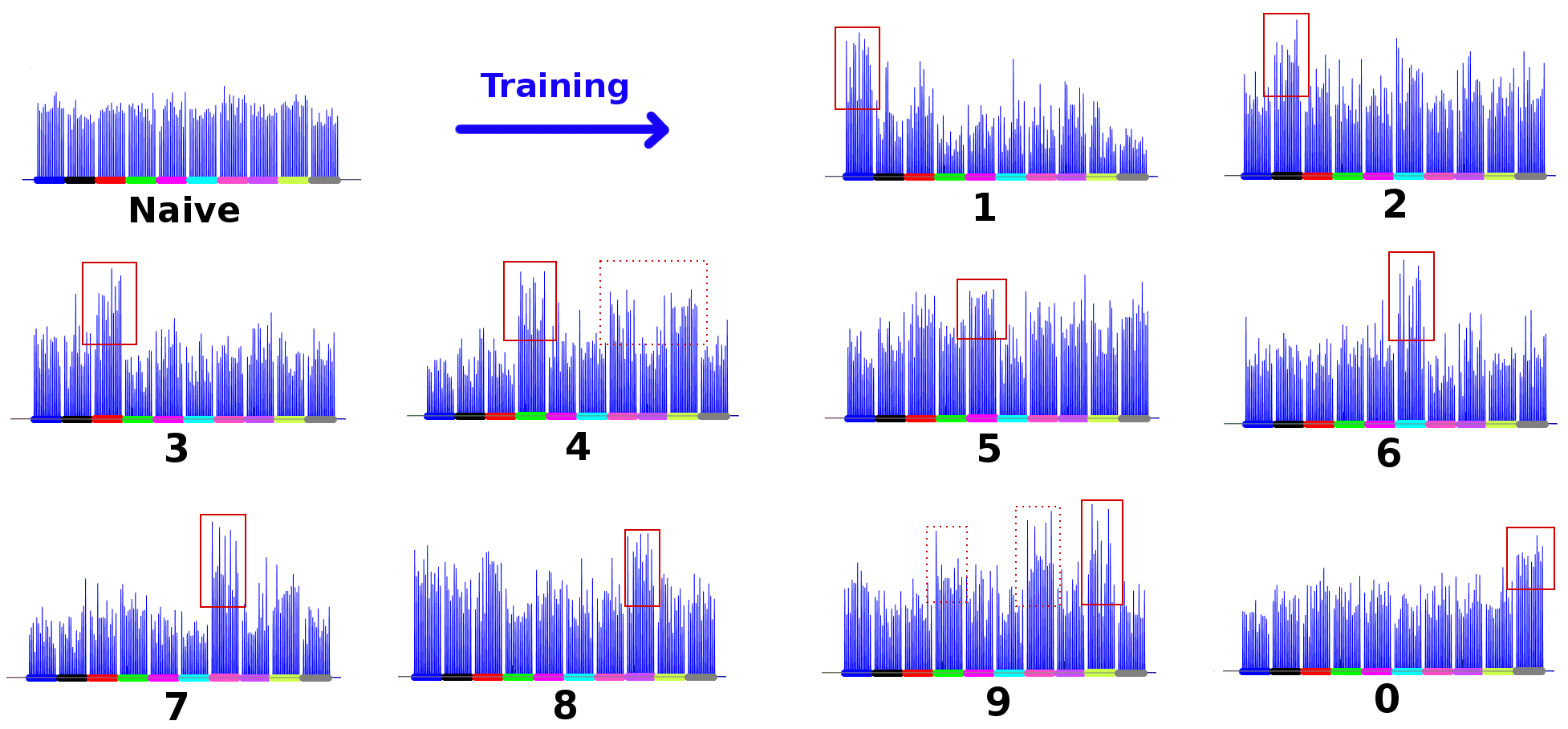} 
	%\rule[.5cm]{0cm}{0cm}
}
\caption{\small {  EN time courses  for a typical MothNet experiment before and after training, showing how training causes  the class response distributions to diverge in each EN.
Each block of spikes shows the time-course of an EN's response to 150 digits (15 ones, then 15 twos, etc). 
The variation in response (spike height) to digits of a single class reflect variations in the digits and also effects of AL noise.
The top left block shows the naive responses (all ENs had similar responses). 
The blue arrow represents the training stage.
The numbered blocks show the responses of the various ENs to 150 validation digits after training has completed.
The EN's responses to digits in the targeted class are framed in red, and some confounding class responses in dashed red). 
This figure shows time-evolutions from one MothNet instance. 
The equivalent statistical results, over several MothNet instances, are shown in Fig \ref{divergingDistStats}.   } }
\label{exampleTimecourse}
\end{figure}

%----------------------------------------------------------------------------------------------------------------------

These behaviors are evident in Fig~\ref{exampleTimecourse}, which shows timecourses of EN firing rate responses, pre- and post-training, for a typical moth. 
Each subplot shows EN response to 150 digits (15 ones, then 15 two's, etc).
The post-training responses are normalized by the EN's mean trained class response for clarity.
Training typically increased and/or decreased all class responses of an EN, but to different degrees, resulting in the separations seen in the normalized timecourses.
Training stage responses are excised from the timecourses to save space; these responses were consistently much stronger due to the stimulating effect of octopamine injected during training, and would extend past the top of the plot. 

\begin{figure}[h!]
\centering
\fbox{\rule[.5cm]{0cm}{0cm} 
\includegraphics  [ width=0.96\linewidth ]{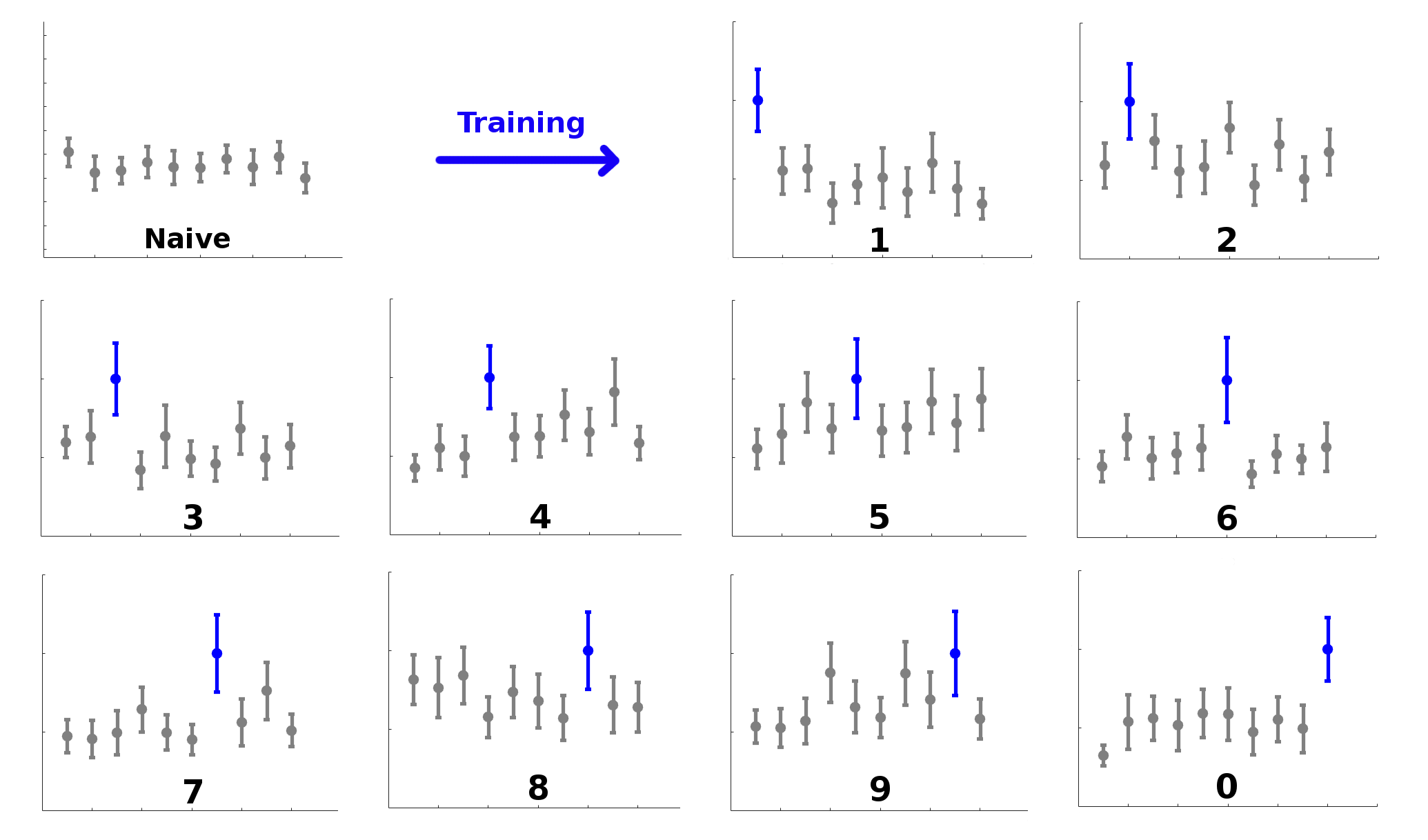} 
\rule[.5cm]{0cm}{0cm}}
\caption{ \small { Statistics of EN response distributions, before and after training, from 13 MothNet instances, showing how training causes  the class response distributions to diverge in each EN. 
The top left subplot shows an EN's naive responses (all ENs had similar responses).  
The blue arrow represents training.\newline
The numbered subplots show the situation post-training:
Let $\mu_{ij}$ be the mean response of the \textit{i\textsuperscript{th}} EN to digits of class $j$ in a single MothNet experiment ($\mu_{ij}$ will be different pre- and post-training). 
Let $\sigma_{ij}$ be the std dev of the \textit{i\textsuperscript{th}}  EN's responses to digits of class $j$ in that same MothNet experiment ($\sigma_{ij}$ will be different pre- and post-training).
Then in the $i^{th}$ numbered subplot, the $j^{th}$ dot gives the mean($\mu_{ij}$), and the $j^{th}$ bar gives $\pm$ mean($\sigma_{ij}$), over post-training $\mu_{ij}$ and $\sigma_{ij}$ from 13 MothNet instances.  
The EN's responses to digits in the targeted class are in blue, responses to the non-targeted classes are in gray.
Mean trained accuracy for this template was 76\%, range 71-83\%. 
This figure is a population-level, statistical analogue to the individual MothNet results shown in Fig \ref{exampleTimecourse}.
} }
\label{divergingDistStats}
\end{figure}

Fig \ref{divergingDistStats} plots EN response distribution statistics (mean $\pm$ std dev) from a typical experiment, in which 13 MothNet instances were generated from a template then trained on 15 samples per class.
Post-training accuracy (log-likelihood) for MothNet instances of this template was 71-83\%,  starting from $\sim$15\%  baseline accuracy.
Similar results, both accuracies and limitations (such as confusing 4s and 9s) held for a wide range of parameter templates and training regimes.
These learning experiments indicate that a model of the moth olfactory network, with minimal modifications, can rapidly learn to read handwritten digits. 
This aptitude was qualified by an apparent upper limit of about 85\% on mean population accuracy (though gifted individuals attained up to 85\% accuracy), given such constraints as the restriction to ``natural'' model parameters, downsampled images, and the use of vectorized (non-spatial) input features.

%--------------------------------------------- 

%

%  \clearpage
%
%----------------------------------------------------------------------------------------------------------------------
%
\subsection{Comparison to standard ML methods}
To set a learning performance baseline, we trained four standard ML methods (Nearest Neighbors, SVM, Neural Net, and CNN) in the limited-training-data regime ($N \leq 100$ samples per class), using the same downsampled, non-spatial dataset (except CNN, see below).
Full hyperparameter details for ML methods and MothNet are given in S.I. and in the online codebase.
We note that while these ML methods can attain over 99\% accuracy on the full MNIST training set (6000 samples per class, with spatial structure) [\cite{leCun95Comparison}], the few-samples regime is fundamentally different, and standard ML methods are evidently not well-suited to it, compared to biological systems.
This few-samples regime  requires that (for MNIST) one ignore 99.9\% of the usual training data.
In addition, the downsampling and vectorization of input images affects accuracy.

Given $N < 10$ training samples per class, MothNet (with log-likelihood classifier) substantially out-performed all the ML methods, and slightly outperformed ML methods at $N$ = 10.
However, MothNet (log-likelihood) appears to have limited capacity, and maxed out at a mean accuracy of $\approx$75\%.
Thus, the ML methods began to pull ahead of MothNet (log-likelihood) at $N$ = 30 to 100, depending on ML method.

\begin{figure}[h!]
\centering
\fbox{\rule[.5cm]{0cm}{0cm} 
\includegraphics  [ width=0.96\linewidth ]{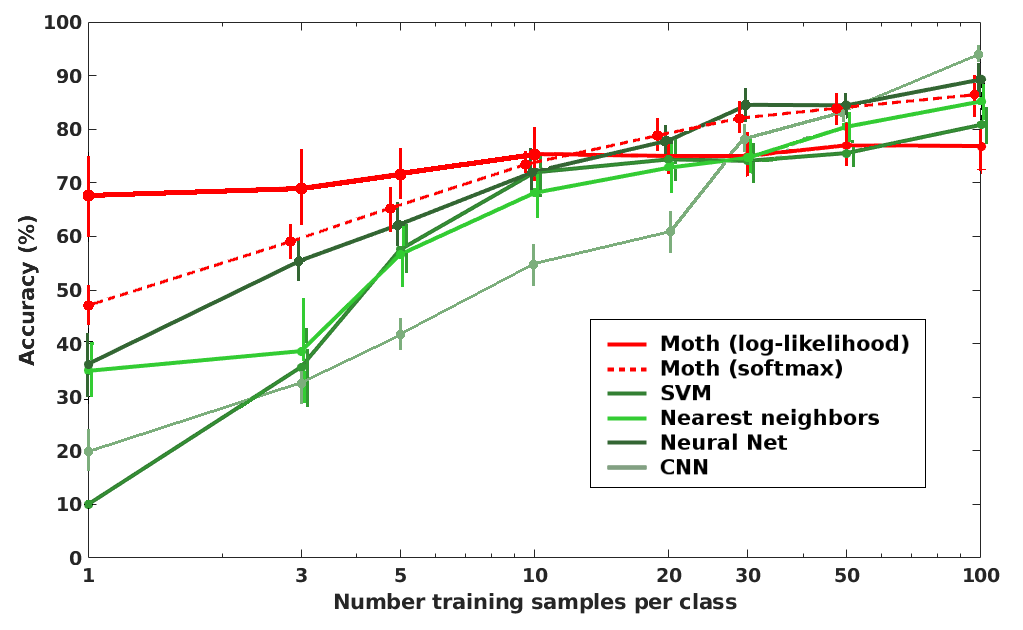} 
\rule[.5cm]{0cm}{0cm}}
\caption{ \small { Bug vs machine: Mean post-training accuracy vs log of $N$ = training samples per class. 
Results are shown for MothNet (in red), as well as for Nearest-Neighbors, SVM, and Neural Nets (all on downsampled  $v$MNIST, i.e. no spatial info); and CNN (on standard MNIST).
At $ N<$ 10 training samples per class), MothNet significantly outperformed standard ML methods. 
Bars show std devs. 
13 runs per data point.
} }
\label{accuracyVariousMethods}
\end{figure}

MothNet (with softmax classifier) had similar accuracy to the NN, outperforming NNs at $N$ = 1 and being roughly equivalent thereafter.
The ML methods never pulled ahead of MothNet (softmax) for $N \leq 100$.
Mean post-training accuracies of the various methods are plotted in Fig~\ref{accuracyVariousMethods}, vs number of training samples per class.

%--------------------------------------------- 

The MothNet accuracies shown in Fig \ref{accuracyVariousMethods} are for MothNet instances randomly generated according to templates from [\cite{delahuntMoth1}], with various learning rates and numbers of ``sniffs'' (see \ref{growthRateSection} and \ref{sniffingSection} for details). 
Nearest-Neighbors and SVM used built-in Matlab functions with z-scaled input features, and with other hyperparameters (number-of-neighbors, box constraint) optimized for each $N$.
The Neural Net was coded in Tensorflow. It had one hidden layer with 85 units (more layers and/or units did not help), Gaussian noise, and dropout.
Learning rate and number of epochs were optimized for each $N$.
For Nearest Neighbors, SVM, and NN, the features were the 85 vectorized pixels  from images pre-processed exactly as for MothNet.

The CNN was a packaged Matlab example, with the number of epochs optimized for each $N$.  
Alone of all the methods, the CNN used standard MNIST (28x28) images with spatial content, on which it can attain $>$ 99\% accuracy given $N \geq 500$. 
Using full-sized images obviated the need to alter the CNN network parameters to fit smaller images, and thus guaranteed a capable architecture.

MothNet with log-likelihood classifier uses statistics from the validation set, as described in \ref{classifier}.
MothNet with softmax classifier does not, in common with the ML methods.

%We note that in our experimental setup, all the methods (except Nearest-neighbors) used the validation set (which doubled as our test set) in some way: 
%The MothNet log-likelihood classifier uses statistics from the validation set, per \ref{classifier}.
%All the ML methods, and MothNet-with-softmax, used results on various validation sets to define key training hyperparameters (except for Nearest-neighbors, since the optimal ``number of neighbors" is obvious in the few-samples regime).

\subsection{One-shot moth}

A variety of ML methods focus on one-shot learning.
These have been tested on Omniglot, an MNIST-like dataset with 1623 alphabetic characters with 20 samples each, created and introduced by [\cite{omniglot}]. 
Some of these ML methods were also applied to MNIST.

We applied MothNet to the Omniglot dataset, and where possible compared it with results from these one-shot ML methods, on both Omniglot and $v$MNIST.

We note that these methods have distinct goals beyond the one-shot learning task. 
[\cite{lake2011}] seeks insight into human transfer learning via extraction of stroke patterns from drawn characters.
[\cite{woodward2017}] explores an active learning system to speed up learning via good training sample choice.
MANN [\cite{santoro2016}] augments a NN with an external memory.
The Matching Network [\cite{vinyals2017}] also adds external memory to a NN or CNN, and has success on complex datasets (e.g. ImageNet).
The Siamese Network [\cite{koch2015}] uses pairs of NNs or CNNs to develop a similarity detector which, tuned by pre-training, can detect similarities in new class instances.
The Neural Statistician [\cite{edwards2016}] learns to compute statistics of new datasets from few samples.
(The Ladder Network [\cite{rasmus2015}] leverages a small labeled training set and a large unlabeled training set from the same classes, a powerful method but for a use-case not treated here.)
This diversity of goals makes comparison between methods inexact and somewhat beside the point. 
We do not wish to imply that a ranking is appropriate or does justice to the various strengths of the methods. 

All comparisons were for one-shot learning. Test datasets included vectorized Omniglot with 20 and with 5 test classes, and $v$MNIST.
Methods that required spatial datasets were omitted.
A major difference between all these ML methods and MothNet is that all the ML methods require pre-training on a large set of samples from separate-but-similar classes (transfer learning) while MothNet assumes no pre-training option.
Pre-training (on 24,000 samples) always used Omniglot data.

Due to the larger size of Omniglot images, the number of input features in MothNet's architecture was increased from 85 to 200, with MB size increased proportionally.
Because Omniglot images are pure binary, they were smeared with a 2-D Gaussian before vectorization and input to MothNet.

Findings of interest include:
\begin{enumerate}
\item (Vectorized Omniglot 20x) MothNet (log-likelihood) surpasses all ML methods, while MothNet (softmax) trailed all ML methods.
Thus, while the MothNet model arguably encodes a stronger classifier, it cannot access its full potential without use of a validation set. 
\item (Vectorized Omniglot 5x) MothNet (log-likelihood) was comparable to ML methods, while MothNet (softmax) accuracy was lower.
For reasons unclear, the difference in MothNet classifiers was smaller on the 5x vs the 20x task. 
\item ($v$MNIST) Since pre-training was on Omniglot characters, the test set in this case did not very closely match the pre-training data. 
This lowered ML method performance by 20\% (for methods using spatial datasets). 
MothNet's accuracy was similar on the Omniglot and $v$MNIST tasks, likely because MothNet did not need pre-training data.
Thus MothNet (log-likelihood) accuracy on $v$MNIST exceeded, and MothNet (softmax) matched the estimated performance of ML methods on $v$MNIST. 
\end{enumerate}
 
A rough summary of this narrow use-case (one-shot learning on these vectorized datasets) might be: 
(i) When the Test set closely matches the pre-training data, MothNet encodes a similar amount of class information as the ML methods, but requires a validation set to fully access it; 
(ii) when the Test set is somewhat distinct from the pre-training data (e.g. $v$MNIST vs Omniglot), MothNet encodes significantly more class information and attains higher accuracy (with both log-likelihood and softmax); and 
(iii) MothNet (uniquely) needs no pre-training. 

\begin{table}[h!]
 \begin{center}
\scriptsize
\caption{Comparison of one-shot methods on vectorized datasets}
\label{table1}
\begin{tabular}{l|c|c|l}
\toprule
\textbf{Method} & \textbf{Test data} & \textbf{\# pre-training } & \textbf{Test Accuracy (\%)  }\\
%\hline
\midrule
%Matching CNN & Omniglot 20x & Spatial & 24k & $~~~~~~~$ 94 \\
%Neural Statistician &  & Spatial & 24k & $~~~~~~~$ 93 \\
%Siamese CNN &  Omniglot 20x & Spatial & 24k &  $~~~~~~~$ 92 \\
Siamese NN &  Omniglot 20x &  24k &  $~~~~~~~$ 58 \\
MANN &  (15x) &  24k &   $~~~~~~~$ 61 \\
MothNet (Softmax) & &  0 &  $~~~~~~~$ 31 \\
MothNet (logLike) &  &  0 &  $~~~~~~~$ {\bf{65}} \\
%\hline 
\midrule
MANN & Ominglot 5x &  24k &  $~~~~~~~$ {\bf{82}} \\
Woodward/Finn &  &  24k &  $~~~~~~~$ 72 (2-shot)\\
MothNet (softmax) &  &  0 &  $~~~~~~~$ 57 \\
MothNet (logLike) &  &  0 & $~~~~~~~$  76\\
%\hline 
\midrule
%Matching CNN && Spatial & 24k &  $~~~~~~~$ 72 \\
%Neural Statistician &    &     Spatial & 24k & $~~~~~~~$ 79 \\
%Siamese CNN & MNIST 10x  & Spatial & 24k &  $~~~~~~~$ 70 \\
Siamese NN &  $v$MNIST 10x &  24k &  $~~~~~~~$ 44 (estimate) \\
MothNet (softmax) &  &  0 &  $~~~~~~~$ 47 \\
MothNet (logLike) &  &  0 &  $~~~~~~~$ {\bf{72}} \\
%\hline
\bottomrule
\end{tabular}
 \end{center}
{\footnotesize{$~$\newline
Notes: Results for ML methods are drawn from their papers, and are restricted to methods that were tested on vectorized datasets. 
Siamese NN performance on $v$MNIST was estimated using Siamese CNN accuracy on spatial MNIST and the ratio of  spatial (CNN) vs vectorized (NN) Siamese accuracies on Omniglot (i.e. $70(\frac{92}{58}$).  
All ML spatial methods had similar drops from spatial Omniglot to spatial MNIST (Siamese CNN,  92$\rightarrow$70\%;
Matching CNN,  94$\rightarrow$72\%; 
Neural Statistician,  93$\rightarrow$79\%).
MothNet results on Omniglot are averages over 11 runs, each with randomly selected target classes.}}
\end{table} 
%  \clearpage
%
%----------------------------------------------------------------------------------------------------------------------------		

\subsection{Growth rate effects, one-shot learning} 
\label{growthRateSection}
The speed at which MothNet learns, i.e. the number of training samples required to reach maximum accuracy, is determined to large degree by the Hebbian growth rate, and the related decay rate, on MB$\rightarrow$EN connections. 
When the connection weights hit the rails of their dynamic range, no further learning is possible (we did not modify the maximum synaptic weight constraints).
We ran experiments to study the effect of Hebbian growth rates on post-training accuracy, given various training set sizes.
One MothNet template (the ``natural learner'') had a biologically plausible growth rate (per \textit{in vivo} data), while the second   template (the ``fast learner'') had a growth rate ``turned up to 11''. 
High growth rate templates allowed us to test MothNet's skill at one-shot learning (i.e. given just one training sample per class).

Fast learners attained strong immediate accuracy (mean 75\%) on just one training sample, but additional training samples provided no further gain. 
This is because the MB$\rightarrow$EN synaptic weights have fixed outer rails: Growth must stop when $w_{ij}$ hits the top rail, and decay must stop when $w_{ij} = 0$. 

Natural learners took several ($\sim$20) training samples to attain maximum accuracy, but that final accuracy was slightly higher.
%When classification was done by simple thresholding (Eqn \ref{thresholdEqn}) rather than the log-likelihood method (Eqn \ref{logLikelihoodEqn}),  the long-term advantage of natural over fast learners was more pronounced (43\% vs 33\%). 
The trade-off of learning speed vs maximum attained accuracy is seen in Fig \ref{fastSlowOneShot}A.
The solid curves show fast- and natural-learner accuracies vs number of training samples.
%The dashed curves show results from the same experiments, using simple thresholding as an alternative classifier.

%Fig \ref{fastSlowOneShot}A incidentally shows the lower accuracy delivered by using thresholds on EN responses (Eqn \ref{thresholdEqn}) versus the log-likelihood classifier (Eqn~\ref{logLikelihoodEqn}). 
%The difference seen here (75\% vs 40\%) was typical.

%
%----------------------------------------------------------------------------------------------------------------------------
%
\subsection{Sniffing, effects of AL noise}
\label{sniffingSection}
\paragraph*{Sniffs}
Biological NNs have a remarkable ability to learn from very few training samples.
In addition, ``sniffing'' behavior (repeated sampling of a given odor) is a common biological strategy [\cite{wilson2008, martin2011}].  
We ran experiments to see whether sniffing behavior improved the various stages of learning.
Sniffing applied to test samples did not improve test accuracy (experiments not shown).

\begin{figure}[t]
\centering
\fbox{  %\rule[ .5cm]{0cm}{0cm} 
\includegraphics [width=0.96\linewidth ] {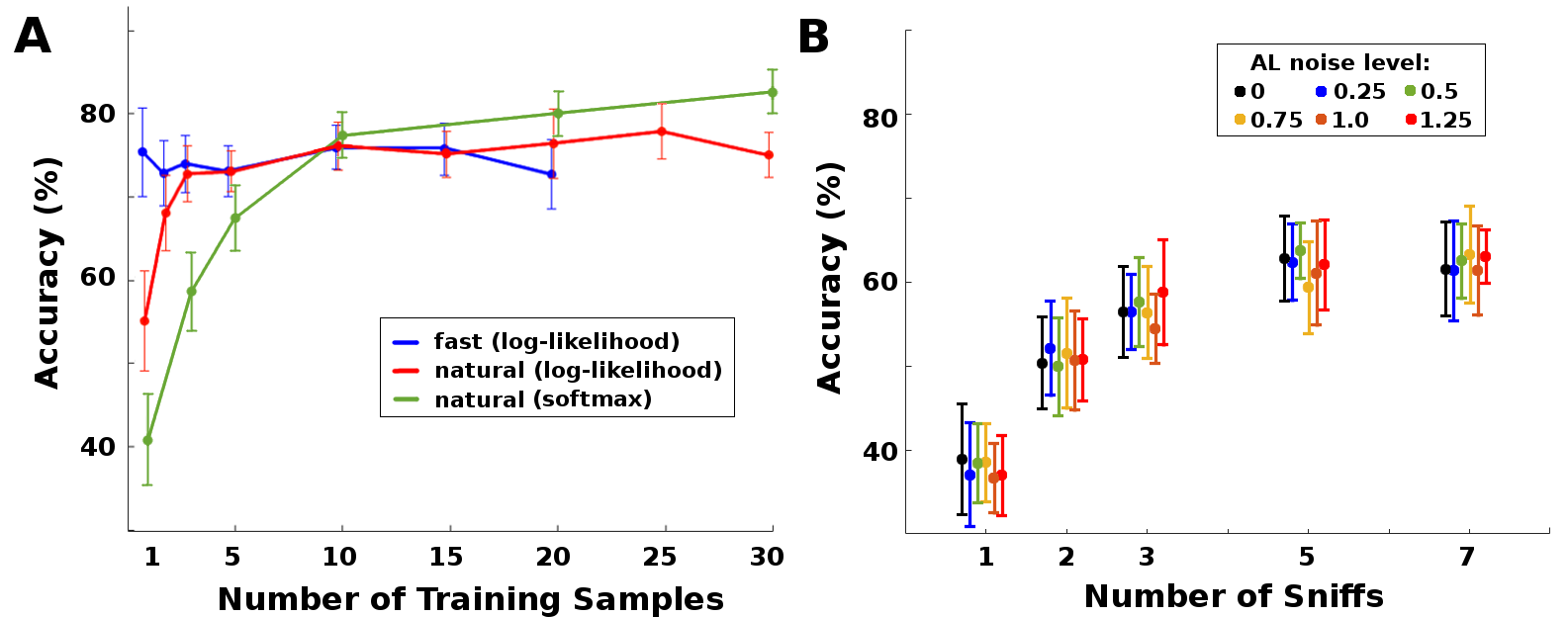}  
%\rule[ 0.5cm]{0cm}{0cm}
} 
\caption{\small{{\bf{A:}} Effects of learning rate parameters. Mean accuracy $\pm$ std dev ($\mu \pm \sigma$) for a  
natural learner (red), a fast learner (blue), and a learner using softmax classifier (green).
For all data points, number of sniffs was varied to maximize trained accuracy.
The fast learner attained 75\% accuracy in one-shot, but with no further gains. 
The natural learner started lower but ultimately attained higher accuracy. 
There was no difference between fast and natural learners using Softmax. 
{\bf{B:}} Effects on one-shot learning of (i) AL noise and (ii) sniffing.
(i) AL noise: Each cluster of $\mu \pm \sigma$ bars represent varying levels of AL noise, with low-to-high noise plotted in different colors from left-to-right at each x-axis location.
AL noise level did not  affect accuracy. 
(ii) Multiple sniffs greatly improved one-shot accuracy. 
Results for a ``natural'' learner with log-likelihood accuracies are shown; Softmax results were similar. 
 13 MothNet instances per data point.
} }
\label{fastSlowOneShot}
\end{figure}

However, sniffing during training had a strong effect, especially in one-shot regimes. 
When a ``natural learner'' (i.e. with learning rate was set to a biologically reasonable level) was given a single training sample (one-shot), multiple sniffs raised post-training accuracy from ~35\% to over 60\%. 
Five sniffs delivered maximal increase, with no further gains from additional sniffs. 
This increase in accuracy due to sniffing in the one-shot context is seen in Fig \ref{fastSlowOneShot}B.
The use of sniffing may partly explain why MothNet was able to reach 80\% accuracy with fewer samples per class (20 vs 200+) than the honeybee MB model in [\cite{huertaNowotny2009}].

\paragraph*{AL noise} 
The moth AL is a very noisy system.
That is, neural responses to a given odor stimulus (or absent any odor) have high variance. 
To see whether this AL noise was beneficial to learning performance, we also varied the AL noise level during the sniffing experiments, from near-zero through the high end of biologically reasonable (per \textit{in vivo} data).
We had expected the presence of AL noise to improve performance, given multiple sniffs, by acting as a kind of sample augmentation, analogous to that used in deep NN training.
In fact, noise levels in the AL had no effect on accuracy.
This is seen in Figure \ref{fastSlowOneShot}B, where each cluster of mean $\pm$ std dev bars shows different levels of AL noise for a given number of sniffs. 

%----------------------------------------------------- 

%
%
%\clearpage

%%----------------------------------------------------------------------------------------------------------------------------
%

\subsection{Sparsity experiments}\label{sparsity}
High-dimensional, sparse neural layers are a widespread motif in biological NNs [\cite{ganguli2012}].
To examine the effects of sparsity in the context of learning, we trained a reasonably capable MothNet template, varying only the sparsity level in the MB during training (17 MothNet instances per sparsity level).
Sparsity here is measured as the fraction of MB neurons that are responsive to stimuli (1\% is very sparse, 50\% is very dense). 
In MothNet, MB sparsity levels (in both non-training and training modes) are parameters.
Sparsity levels in the MB affected two crucial behaviors:  Intra-class signal-to-noise ratio (SNR) of EN responses; and how well a given EN$_j$'s response to training focused on class $j$ (learning focus).

Results show that the sparse MB layer plays a key role in learning by controlling and focusing Hebbian weight updates.

High MB response fraction (i.e. high density, or low sparsity) correlated strongly with high SNR (i.e. reliability of intra-class EN responses).
But it also resulted in poor post-training classifier accuracy, because Hebbian growth was not focused:
Rather, it boosted all weights due to the excess of active MB neurons, so weight increases into a particular EN were not restricted to just the most class-relevant signals.
The trained EN then responded strongly to all digits rather than just its target digit.

Conversely, low MB response (i.e. high sparsity)  resulted in low intra-class SNR, since not enough MB neurons would fire in response to a digit for reliable activation of the ENs.
But high sparsity correlated strongly with high ``learning focus'':
Training focused gains on the correct class, ensuring that ENs' responses to their targeted digits were preferentially strengthened. 
This resulted in stronger post-training accuracy. 

Post-training discrimination accuracy appeared to represent a tradeoff between intra-class SNR and learning focus.
This tradeoff is shown in Fig \ref{sparsity}, which plots these effects of different sparsity levels in the MB, as they relate to learning: 
%\begin{enumerate}
%\item 
\paragraph*{Descending red curve} This plots ``learning focus'', a figure-of-merit defined as the average standardized distance between EN response distributions to trained and control classes,
\begin{equation} 
LF = \frac{1}{9}\sum\limits_{i\neq j} \frac{ (\mu E_{jj} - \mu E_{ji}) } { 0.5(\sigma E_{jj} + \sigma E_{ji} ) } \text{, notation as in Eqn~\ref{logLikelihoodEqn}.}
\end{equation}
This is average Bhattacharyya distance if EN response distributions are gaussian.
In very sparse regimes training was strongly focused on the trained class, while in dense regimes training ``raised all boats'' and the trained response distributions were poorly separated.
Thus high sparsity focused learning well, and low sparsity diluted it. 
This finding matches results in [\cite{huertaNowotny2009, peng2017}]. 
%\item  
\paragraph*{Ascending black dotted curve} This plots the mean intra-class signal-to-noise ratio (SNR):
\begin{equation} \label{snrEqn}
\text{SNR} = \frac{\mu(f)} {\sigma(f)} \text{ where } f  = \text{ EN odor response;}
\end{equation} 
This was an opposite situation: A very sparse MB resulted in high intra-class variance in trained EN responses (low SNR), while a denser MB delivered much more consistent within-class responses (high SNR).  
%\item 
\paragraph{Domed blue curve} This is a fit to post-training classifier accuracies (a natural objective function) from experiment.
%\end{enumerate}

Judged by  best post-training classification accuracy,  optimal MB sparsity level for MothNet  appears to represent a compromise between delivering sufficient intra-class SNR and sufficient ``learning focus" for inter-class distinctions.
This optimal region lies somewhere between 5-20\%, consistent with the 5-15\% commonly observed in live BNNs.

%---------------------------------------------------------- 

%
\begin{figure}[t]
\centering
\fbox {  %\rule[ .5cm]{0cm}{0cm} 
\includegraphics [width=0.96\linewidth ] {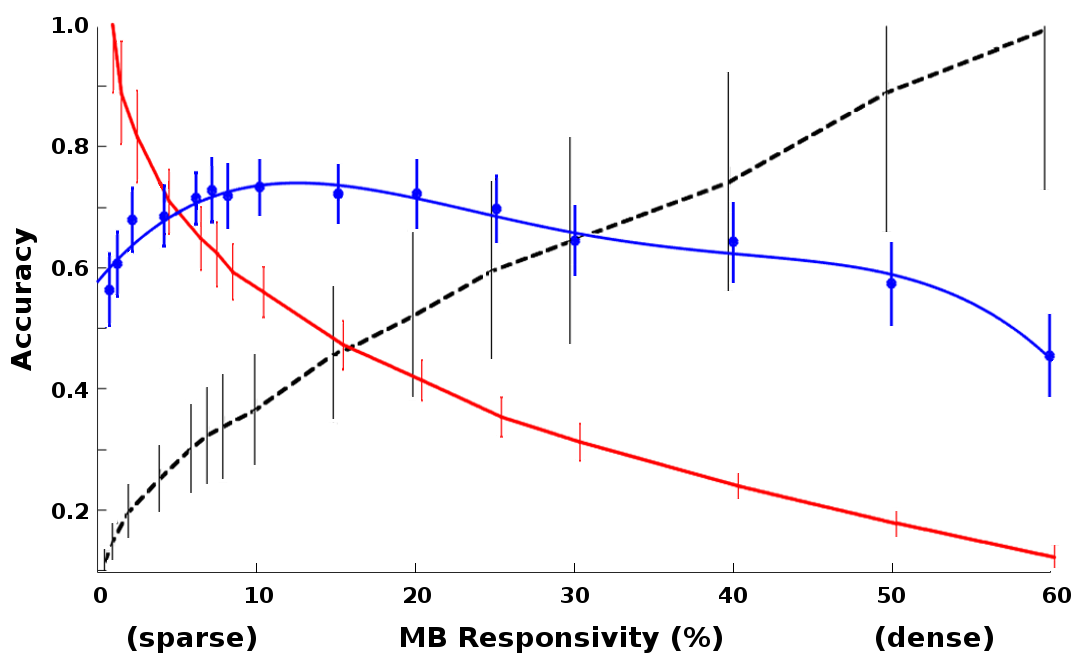}  
%\rule[ 0.5cm]{0cm}{0cm}
} 
\caption{ \small { {\bf{Effects of MB sparsity on learning:}}
Blue domed curve = a fit to mean trained accuracies, which peaked at 5-20\% sparsity, a compromise between learning focus and high intra-class signal-to-noise ratio (SNR).
Means $\pm$ std devs ($\mu \pm \sigma$) are shown.   
Descending red curve = $\mu \pm \sigma$ of learning focus (separation of responses to trained vs control digits). 
Black ascending curve = $\mu \pm \sigma$ intra-class SNR. 
Learning focus and SNR are scaled for plotting.
17 MothNet instances per sparsity level.
} }
\label{sparsity}
\end{figure} 
%
%\clearpage

%--------------------------------------------------------------------------------------------------------------------------------
	
%--------------------------------------------------------------------------------------------------------------------------------------------------------------
%------------------------------------------------         DISCUSSION               ------------------------------------------------------------------------
%--------------------------------------------------------------------------------------------------------------------------------------------------------------

%
\section{Discussion}
\subsection{A biological toolkit for building NNs}

Our goal has been to provide a proof-of-concept as to the learning abilities of even the simplest of biological neural networks (BNNs), and to clarify how a NN built from biological elements learns a predictive model for a general ML task.

In order to learn new odors, the moth olfactory network uses just a few core tools: A noisy pre-processing network with competitive inhibition; Hebbian plasticity controlled by a high-dimensional sparse layer; and generalized (global) stimulation during training.

Our key finding is that a neural net built with this biological toolkit can succeed at a general rapid-learning task, and in fact can out-perform standard ML methods. 
MothNet learned to read $v$MNIST digits, increasing its accuracy more than 5x (from 15\% to 75-80\%) given only a few training samples (1-20 per class).
Also, because Hebbian weight updates focus entirely on activity induced by the class being trained (not on activity induced by control classes, as in backprop), new classes can be added and trained without retraining on existing classes.

This finding is of interest because the biological tools analyzed here are well-suited to being combined and stacked into larger, deeper neural nets, just as convolutional kernels, maxpool, etc, are combined to build current DNNs.
The success of live BNNs  at a wide range of tasks argues for the potential of NNs built with a biological toolkit to succeed at ML tasks.
%They can learn from few samples (even just one), and can expand to include new classes without retraining, both areas of weakness for current DNNs.  

We note that in these experiments MothNet stayed close to the moth's very simple olfactory architecture, and used a simple feature set as input (pixels). 
We believe that a free hand with network design, such as using better input features, biologically-unrealistic model parameters, and more complex architectures, would yield stronger results in a variety of tasks.
This effect has been demonstrated by biological brains of increased size and complexity, and indeed by increasingly complex DNNs.

\subsection{The rapid learning regime}

It is not, on reflection, surprising that an insect brain might out-perform ML methods at a few-samples learning task.
First,  BNNs in general excel at rapid learning from few samples. 
Second, in the typical ML context, such as a competition or a comparison with other methods using well-defined benchmarks, the number of training samples is fixed and is often high (e.g. 60,000 for MNIST). 
The competitive pressure is to maximize accuracy, with no penalty for using lots of training data.
In sharp contrast, an insect pays a high cost for every extra training datum, and their competitive pressure is to very rapidly attain ``good-enough'' accuracy.
 
These two regimes complement each other, and there are applications where it may be advantageous to pair a fast, rough learner with a slower, more precise learner.
One example might be an adaptive controls learner for a drone: 
If the drone collides in mid-air and loses an engine, you eventually want to learn a new, optimal control strategy for the reduced system.
But the first necessity is to learn a ``good enough" control strategy very fast, before crashing, to allow time for the subtler system to train.
Another example might be a multi-stage bootstrap labeling scheme when labeled data are scarce: 
In the initial stage, given just one or a few labeled samples, the fast learner might classify and apply labels to more samples (perhaps keeping only those with the most certain labels) until there are enough labeled samples to train a subtler system as the second stage labeler.
%
%----------------------------------------------------------------------------------------------------------------------------
%

\subsection{Role of the Hebbian update rule}

We hypothesize that the ability of BNNs such as MothNet to generalize well from very few samples is related to the Hebbian update mechanism, which operates differently than statistically-based optimizers. 
It has no objective function or output-based loss that is pushed back through the network  as in backprop or agent-based reinforcement learning.
Rather, weight updates occur on a purely local ``use it or lose it'' basis.
Typical ML optimizers, such as backprop, excel at interpolation [\cite{mallat2016}], but arguably not extrapolation. 
Perhaps Hebbian update rules, which strengthen or decay connection weights according to amount of traffic during learning episodes, enable effective generalization outside the convex hull of interpolation. 
%
%----------------------------------------------------------------------------------------------------------------------------
%

\subsection{Role of the sparse layer} 
A second finding emphasizes the key role of sparse layers during learning. 
Sparse, high-dimensional layers are a widespread motif in biological neural systems, especially in networks related to memory and plasticity [\cite{ganguli2012}].
In the MothNet model, the sparse layer (MB) plays a vital role in learning because all the plastic synapses connect into or out of the sparse layer, allowing it to modulate the Hebbian updates to the synaptic connections by taking advantage of the fact that Hebbian growth is an AND gate (``fire-together, wire-together'') [\cite{delahuntMoth1}].
This ensures that learning boosts the important signal (i.e. the signal associated with a given sample) and not artifacts.

While the sparse MB layer calls to mind the backprop sparse autoencoder [\cite{ngSparseAutoencoder}], the biological role described here has no obvious analogue in backprop sparse encoders, since it is tied to the Hebbian update method. 
However, biological sparse layers may also perform functions similar to those found in backprop sparse autoencoders, such as reducing noise [\cite{vincent2008}] and reducing the dimension of the feature space in order to better match the essential dimension of the classification task [\cite{makhzani}].
%
%----------------------------------------------------------------------------------------------------------------------------
%

\subsection{Role of competitive inhibition}
A key feature of the AL layer is competitive inhibition between units, each of which receives input from exactly one feature.
This competitive inhibition may enable rapid learning by creating several attractor basins for inputs, each focused on a particular class according to which features present most strongly.
This might serve to push otherwise similar samples (of different classes) away from each other, towards their respective class attractors, increasing the effective distance between the samples. 
Thus the  outputs of the AL, after this competitive inhibition, may have better separation by class.
The fact that the MothNet model includes an AL may be one reason  it was able to learn more rapidly than the honeybee MB-only model in [\cite{huertaNowotny2009}].
%
%----------------------------------------------------------------------------------------------------------------------------
%

\subsection{Role of octopamine} 
Unanswered by these experiments is whether generalized stimulation by octopamine is required in ML systems such as MothNet, distinct from actual biological systems. 
In the moth, octopamine stimulation may offer a work-around to avoid biological constraints on Hebbian growth rates and input intensity. 
That is, octopamine may act primarily as an accelerant. 
However, engineered NNs can easily crank up growth rates, enforce higher MB activity, and amplify signals during
training, all without recourse to the octopamine mechanism but perhaps with the same beneficial effect on learning. 
In this case, generalized stimulation would not be a necessary part of the biological toolkit described here for application to ML tasks. 
However, it may be that octopamine stimulation also enables exploration of the coding solution space not normally activated by stimuli.
Alternate ways to replace this functionality in the ML context are not so obvious.

%
%----------------------------------------------------------------------------------------------------------------------------
%

\subsection{Role of noise} 
Also unclear is the role during learning, if any, of the high intrinsic noise built into MothNet's Antennal Lobe.
In a biological context neural noise (including in the AL) can have diverse benefits independent of learning [\cite{knight1972,wiesenfeld1995,maPougetBayesianInference,delahuntMothInjury}]. 
However, our experiments found no positive reason to build high intrinsic noise into the AL (or any NN layer) in an ML context. 
However, a noisy pre-amp (e.g. the AL) might still be directly beneficial to learning, despite our results, for three reasons.

First, when coupled with sniffing, a noisy AL might provide a version of data augmentation (as used in DNNs) by distorting the codes delivered to the MB and readout neurons. 
This would improve one-shot or few-shot learning, if the distortions induced by the noisy AL somewhat mimicked the within-class sample variation (this condition was likely absent in our MNIST experiments).
Second, injecting noise into input layers, or corrupting training samples, can improve NN classification performance  [\cite{guozhong}], suggesting a concrete benefit during training (though perhaps not during classification). 
Third, injecting noise may be a useful or even necessary way to explore the solution space  [\cite{bengioNIPS2015}].
 %
%----------------------------------------------------------------------------------------------------------------------------
%

\subsection{Expanding the biological toolkit}
The set of algorithmic tools for learning characterized here derive from one of the simplest biological learning networks (a bug brain).
It is still unclear whether modifications to this network can substantially improve performance, or whether this particular architecture  can achieve just so much in terms of accuracy and scope of work, commensurate with the constrained needs and pressures experienced by insects as they evolved.
Happily, BNNs are incredibly diverse and complex.
Countless other useful elements can likely be abstracted from other biological systems and applied to ML tasks.
This requires accurately modeling these more complex systems and analyzing how they learn, an open-ended task. 

\section*{Acknowledgements}
\noindent Thanks to: \newline
$~~~~$Christian Szegedy for valuable draft-editing input; \newline 
$~~~~$Satpreet Singh for insights about the role of the AL;\newline 
$~~~~$The referees, for highly germane feedback and insights.\newline\newline
This work was partially supported by the Swartz Foundation.\newline 

%---------------------------------------------------------------------------------------------------------------------------------
%----------------------------------------------  S. I. -----------------------------------------------------------------------------
%-------------------------------------------------------------------------------------------------------------------------------

\section{Supplemental Information for ``Putting a bug in ML''}
  
This document gives details of the various ML methods used in the paper ```Putting a bug in machine learning''. 
Full code for all methods is available at \url{https://github.com/charlesDelahunt/PuttingABugInML}.

All code was in Matlab, except for the Neural Net (in python).
$N$ refers to the number of training samples per class.
 
\subsection*{Data preprocessing}
All methods (MothNet and ML) were applied to the same pre-processed MNIST data. 
Pre-processing steps were designed to reduce the number of pixels (ie features) towards 60, to match the biological moth's architecture. 
Pre-processing steps included cropping, downsampling, selecting a receptive field of the most active pixels (on a population level), vectorizing, and scaling.
Pixel values were scaled in one of two possible ways:  (i) their sum was set  to a constant (= 6); or (ii) the 90\textsuperscript{th} percentile pixel value was set to a constant (= 0.5).
Pre-processing reduced the MNIST thumbnails to 85 vectorized pixels-as-features.

The CNN model, exceptionally, used full-sized MNIST images with spatial order retained.

When MothNet was applied to the Omniglot dataset, the thumbnails (which are binary images) were smeared with a 2-D gaussian prior to pre-processing and vectorizing as in MNIST.

Results for one-shot ML methods came from their papers, so pre-processing (of both Omniglot and MNIST) was defined by the respective methods. 
In general they involved downsampling, and sometimes vectorization.

\subsection{MothNet} 

For MNIST, MothNet had 85 units in the AL, corresponding to 85 features. 
For Omniglot, MothNet had 200 units in the AL (because Omniglot images are 105 x 105 rather than 28 x 28).
In each case, number of MB units was 30 times the number of AL units.

Synaptic weights were drawn from random distributions whose hyperparameters were carried over from ``Biological Mechanisms for Learning'', where they were chosen so that MothNet's neural firing rates accurately matched \textit{in vivo} firing rate data recorded in moth Antennal Lobe neurons.
Each MothNet instance started with random weights prior to training (as in standard NNs).
On Omniglot tests, this approach contrasted with the ML one-shot learners, which required pre-training on similar-but-distinct data to pre-set network weights (no judgement is implied as to which if any method is better).
 
Speed of MothNet learning was controlled in two ways: 
(i) By learning rates, both Hebbian growth and decay, controlled by the parameter ``goal'', corresponding roughly to the number of training sample exposures to attain optimal performance, ie high goal would have lower learning rates; and
(ii) by number of sniffs, ie how many times each training sample was inputted to the network during training.

Pixel value scaling by sum was used for the log-likelihood classifier, while scaling by 90\textsuperscript{th} percentile pixel was use for the softmax classifier.

\subsection{Nearest Neighbors}

For Nearest-Neighbors, we used Matlab's ``fitcknn" function.
Distance was the usual $L_2$ norm on features (pixels).
We  z-scaled the features (option ``standardized'') since this worked better than not z-scaling.
Number-of-neighbors was optimized by parameter sweep for each number of training samples per class.
This gave number-of-neighbors = 1 ($N \leq 10$) or 3 ($N \geq 20$).
Pixel normalization method had no effect on accuracy.

\subsection*{SVM}

For Support Vector Machine (SVM), we used  Matlab's ``fitcecoc" function.
We  z-scaled the features (option ``standardized'') since this worked better than not z-scaling.
We omitted the gaussian kernel option because it gave worse results.
For Matlab's SVM function, the key parameter is ``box constraint'' (higher box-constraint means a higher cost of misclassified samples), which was optimized by parameter sweep for each $N$.
We used the following optimized ``box constraint'' values:
1e4 for $N=3$;  1e0 or 1e1  for $N=5$;  1e-1  for $N=10$,  1e-4 or 1e-5 for $N=20$, 1e-5  for $N=50$; 1e-7  for $N=100$.
Pixel normalization method had no effect on accuracy.

\subsection{CNN}

The CNN was a packaged Matlab example  run from  the Matlab command line with ``openExample('nnet/TrainABasicConvolutionalNeuralNetworkFor\\ClassificationExample')".
Number of epochs was optimized for each $N$ by parameter sweep. 
Alone of all the methods, the CNN used standard MNIST (28x28) thumbnails with spatial content, on which it can attain $>$ 99\% accuracy given $N \geq 500$.
Initial learning rate was set to 0.01.
Layers were:\\
$~~~$conv2d (3x3, 8 filters, batch normalization, ReLU)\\
$~~~$maxPool( 2x2, stride = 2)\\
$~~~$conv2D (3x3, 16 filters, batch normalization, ReLU)\\
$~~~$maxPool(2x2, stride = 2)\\
$~~~$conv2D (3x3, 32 filters, batch normalization, ReLU)\\
$~~~$dense layer(10 units)\\
$~~~$softmax layer

\subsection{Neural Net}

The NN was coded in Tensorflow, and had one hidden layer with 85 units.
Increasing the number of layers, or the number of units per layer, did not yield any performance gains.
Thus, we believe that the NN's capacity was optimized, for this task, by a single layer.
Learning rate was set to 0.01, scaling factor for added gaussian noise to 0.01, each optimized by parameter sweep.
Dropout rate was 0.3 (applied before and after the hidden layer).
Number of epochs  to optimized for each $N$ by sweep. 
The NN was only run on the data scaled by pixel sum.
We thank Christian Szegedy for help with the initial format of the NN, though we emphasize that any failures of optimization are on us.

\section*{References}

%\section{}
%\label{}


\bibliographystyle{elsarticle-harv}   % specified by Elsevier

 \bibliography{mothBibliography_jan2019}

\end{document}